\documentclass[twocolumn]{article}
\usepackage{flafter}
\usepackage[dvips]{epsfig}

\newtheorem{example}{Example}
\begin{document}

\title{Transforming and Enriching Documents for the Semantic Web}

\author{
Dietmar R\"osner,  Manuela Kunze, Sylke Kr\"otzsch\\
Institut f\"ur Wissens- und Sprachverarbeitung\\
Otto-von-Guericke-Universit\"at Magdeburg\\
P.O.Box 4120, 39016 Magdeburg, Germany\\
{{\tt \{roesner$\mid$makunze$\mid$kroetz\}@iws.cs.uni-magdeburg.de}}}

\maketitle

\begin{abstract}
We suggest to employ techniques from Natural Language Processing
(NLP) and Knowledge Representation (KR) to transform existing
documents into documents amenable for the Semantic Web. Semantic Web
documents have at least part of their semantics and pragmatics marked
up explicitly in both a machine processable as well as human readable
manner. XML and its related standards (XSLT, RDF, Topic Maps etc.)
are the unifying platform for the tools and methodologies developed
for different application scenarios.

%
\end{abstract}
\section{Motivation}
\begin{figure}[th]
\centerline{\psfig{file=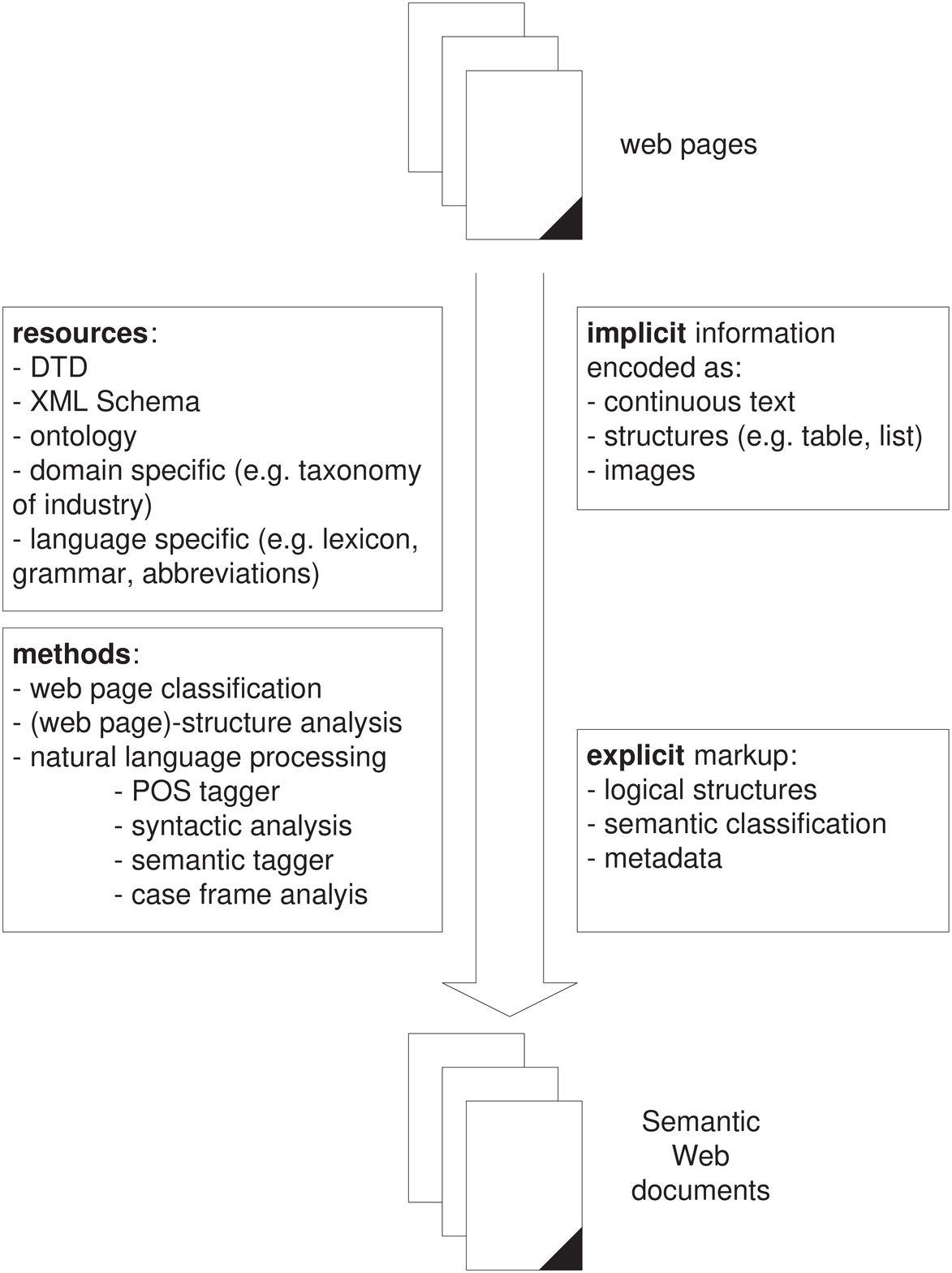,width=3in,angle=0}}
\vspace*{12pt} \caption{Towards the Semantic Web. \label{project}}
\end{figure}

Imagine the following situation: As a consumer you are looking for
information about a product. You may be interested in technical
details, the price, delivery conditions etc. For many products this
type of information is available on web pages of companies.\\
A situation that - despite the differences - is very much alike: As a
scheduler of an automobile \hyphenation{manu-facturer}manufacturer
you are looking for subsidiary companies that are able to produce
components or raw products for integration in a new production
pipeline. Again you might profit from the information that is offered
on web pages. An
illustrative case are web pages of foundries.\\
The problem nowadays still is: Although information searched for
is available on WWW pages automated processes (`agents') will have
problems in finding and extracting it. \\This motivates the vision
of the `Semantic Web' as expressed by Tim Berners-Lee: `Semantic
Web -- a web of data that can be processed
directly or indirectly by machines' \cite{berners-lee:1999weavingweb}.\\
The Semantic Web of the future will very likely be based on direct
authoring \cite{xmldeu2001}. That means future documents will
contain metadata, semantic tagging will be employed to make
intra-document relations explicit, topic maps and other
technologies will be used to express semantic relations between
documents (inter-document relations).  In other words: making
semantics and pragmatics of documents explicit via tagging will be
an integral part of the document creation
process. \\
In the current situation we have a multitude of existing web pages
with valuable contents, far too many to be manually augmented and
transformed into Semantic Web documents. We therefore suggest both
automatic and semi-automatic augmentation of documents.\\
We are developing tools and methodologies based on NLP techniques,
text technology and knowledge representation for the
transformation
of existing documents into Semantic Web documents (see Fig. \ref{project}).\\
In the following we suggest to distinguish two types of documents. On
the one hand there are {\it enriched documents}: These are documents
that originate from the web or other sources. They undergo document
analysis and the results of the analysis are directly integrated into
the document using XML markup.\\ On the other hand there are {\it
transformed documents} that are comprised from explicitly understood
pieces of information extracted from other documents.\\ On the
surface enriched documents may have the same `look and feel' as
simple HTML pages, i.e., on a first glance the user may not recognize
any difference. The added value of the explicit enrichment with
structural and semantic markup becomes apparent when automated
processes (agents) are used. Enriched documents are suited for
intelligent searches, querying, flexible recombination etc.\\ The
case of document transformation comes into play when pieces of
information are reassembled and offered to the user in a uniform way
as a document. As an example you can think of a rated overview page
that is distilled from a collection of pages, e.g., a structured
summary with the results of a comparative search on a
collection of web pages from different manufacturers of a product.\\
A general point to stress: In some sense the terms
`semi-structured documents' or `unstructured documents' that are
used sometimes are misleading from our point of view. Documents
are generally highly structured. The problem is not a lack of
structure, the problem is that the structure is not made explicit
in traditional documents. Human readers are in most cases able to
easily uncover those implicit structures. Thus the major challenge
is for automatic conversion of traditional (web) documents into
Semantic Web documents: to uncover structure and contents and mark
them up explicitly. This will then
allow processing und interpretation of the documents by machines.\\
This paper is organized as follows: first we give a detailed
description of our view of Semantic Web documents. Then we
describe encoding of information on web pages and mechanisms for
transformation of implicit information into explicit data,
followed by a realistic application scenario. Finally we discuss
some of the open research issues.

\section{Towards Semantic Web Documents}
The WWW is a fast growing source of heterogeneous information. For
web document analysis in general, the analysis of text will have to
be complemented by the analysis of other WWW media types: image
analysis, video interpretation, voice processing etc. In addition,
cross-media references and hypermedia structures need proper
treatment.\\ Natural language analysis of textual parts of web
documents is no different from `normal' text
analysis.\\
For a given complex application in web document analysis we found
it fruitful to classify its subtasks into the following three
categories:
\begin{itemize}
    \item subtasks that are primarily WWW specific,
    \item subtasks that are specific to the application,
    \item subtasks that are relevant to all NLP approaches.
\end{itemize}
WWW specific subtasks can be classified as being part of the
preprocessing stage. Preprocessing in this sense comprises all
those operations that eventually result in a text document in the
input format expected by the linguistic tools. In other words,
aspects of the source document that are irrelevant or distracting
for linguistic processing will be abstracted away during
preprocessing and the resulting document will be in a canonical
format.\\
If source documents already contain appropriate metadata some
subtasks of preprocessing are reduced to looking up the values of
metadata attributes. For now, preprocessing will in many cases
include attempts at automatic language identification, domain
classification or hyperlink tracing.

\subsection{The Power of Markup}
XML \cite{bray.paoli.sperberg-mcqueen.maler:2000} -- and its
precursor SGML -- offer a formalism to annotate pieces of (natural
language) text. To be more precise, if a piece of text is (as a
simple first approximation) seen as a sequence of characters
(alphabetic and \hyphenation{white-space} whitespace characters) then
XML allows to associate arbitrary markup with arbitrary subsequences
of {\em contiguous} characters.  Many linguistic units of interest
are represented by strings of contiguous characters (e.g., words,
phrases, clauses etc.). It is a straightforward idea to use XML to
encode information about such a substring of a text interpreted as a
meaningful linguistic unit and to associate this information directly
with the occurrence of the unit in the text. The basic idea of
annotation is further supported by XML's wellformedness demand, i.e.,
XML elements have to be properly nested. This is fully concordant
with standard linguistic practice: complex structures are made up
from simpler structures covering substrings of the full string in a
nested way.
The enrichment of documents is based on this ability to associate
information directly with the respective span of text.

\section{Information Encoding on WWW Pages}
The starting point of our work are web pages as they are found now.
In the following we report about results from corpus based
case studies.\\
\subsection{Corpus Based Case Studies}
We employ a corpus based approach. Investigations do start with
the collection of a corpus of representative documents.\\Then a
number of issues are systematically investigated:
\begin{itemize}
\item What are the typical structure and contents of document
instances in the corpus? \item What information is most likely of
interest for which type of users? \item How can this information be
located and extracted? \item What are characteristics of the source
documents that may make this task easier or more complicated? \item
What aspects can be generalized and abstracted from the specific
case?
\end{itemize}
\subsection{Variations in Information Presentations}
When the focus is on contents, not on surface appearance, then the
notion of `paraphrase' is no longer restricted to linguistic
units. Our analyses of WWW pages revealed that there are many
cases where the same information can more or less be conveyed both
in a number of linguistic variations as well as in
different non-linguistics formats (`extended paraphrases').\\
As an example we take the following excerpt from a web page of a
garage manufacturer.
\begin{example} Excerpt from Web Page. \label{ex_wunschbox}
\small \begin{quote}
Wunschbox-Garagen sind als Typ S mit einer Breite von 2,68m, als
Typ N (Breite 2,85m) und als Typ B (Breite 2,98m) lieferbar. Alle
Garagen haben eine Hoehe von 2,46m.
\end{quote}
\end{example}
\normalsize The phrasal pattern underlying the first sentence
\small
\begin{quote}
\flushleft{\emph{$<$product$>$ is\underline{ }available\underline{
}as $<$enumeration of type info$>$}}
\end{quote}\normalsize
is found in variations like the following: \small\begin{quote}
\flushleft{{\it $<$product$>_{(pl)}$ sind als $<$enumeration$>$ lieferbar.}\\
{\it $<$product$>_{(sg)}$ ist als $<$enumeration$>$ erh\"altlich.}\\
{\it $<$product$>_{(sg)}$ gibt es als $<$enumeration$>$.}
\ldots}
\end{quote} \normalsize
Type info in turn is given
according to patterns like: \small\begin{quote}
$<$enumeration of type info$>$ ==\\
$<$type nr$_{1}$$>$ with $<$feature$>$  of $<$value$_{1}$$>$, \\
      $<$type nr$_{2}$$>$ with $<$feature$>$  of $<$value$_{2}$$>$,\\
     \ldots \\
     $<$type nr$_{i}$$>$ with $<$feature$>$  of $<$value$_{i}$$>$
\end{quote}
\normalsize Note that the second sentence of example
\ref{ex_wunschbox} needs contextual interpretation because its
literal meaning in isolation would be the universally quantified
assertion that `\emph{All garages have a height of 2,46m}' and not
the contextually restricted `\emph{All garages of types S, N
and B of this manufacturer have a height of 2,46m}'. \\
Essentially the same information could as well be given in a
variety of tabular formats.
\begin{table}[h]
\caption{Tabular Information Presentation.}\label{tab_rel} Garage
types:
  \begin{center}
\begin{tabular}{ |r|c|c| }
 \hline
  type & width & height \\ \hline \hline
  S & 2,68 & 2,46 \\ \hline
  N & 2,85 & 2,46 \\ \hline
  B & 2,98 & 2,46 \\ \hline
  \end{tabular}
\end{center}
\end{table}
The third column of table \ref{tab_rel} (height) could be omitted and
replaced by the sentence: {\it All (our) garages have a height of 2,46m.}\\
In general, combinations of linguistic units and tables are
possible, e.g., like in {\it All data of the following table refer
to garage type N.}
\subsection{Interaction Between Linguistic Structures and Source
Document Markup} There are interactions between list structures in
HTML (and XHTML) and linguistic units that have to be accounted for
in linguistic analysis of web pages.\\ A simple case is the use of
lists for enumerating concepts like in the following example.
\begin{example} Enumerating Concepts. \label{ex_enum-concept}
\small
\begin{quote}
\flushleft{$<$p$>$Die wichtigsten Branchen sind: $<$ul$>$
  $<$li$>$Formen- und Werkzeugbau$<$$/$li$>$
  $<$li$>$Eisenbahnwesen$<$$/$li$>$
$<$$/$ul$>$}
\end{quote}
\end{example}
\normalsize Please note that even such simple structures do need
proper treatment of coordination and truncation and that
contextual interpretation is obligatory in order to correctly
infer semantic
relations between list heading and list items.\\
Very often a partial sentence and an HTML list of sentential
complements or other phrases interact like in the following example:
\begin{example} Interaction Between Sentence and List Structure. \label{ex_lists}
\small
\begin{quote}
\flushleft{Wir produzieren maschinen- und handgeformten Grau- und
Sphaeroguss\\ $<$ul$>$\\
 $<$li$>$ca. 25.000 t Jahresproduktion\\
 $<$li$>$mit mehr als 4.000 lebenden Modellen\\
 $<$li$>$in mittleren und groesseren Serien\\
 $<$li$>$Handformguss bis 800 kg Stueckgewicht\\
$<$$/$ul$>$}
\end{quote}
\end{example}
\normalsize Here semantic interpretation unavoidably needs
sophisticated techniques for the interaction between the list
items as phrasal type syntactic structures and the partial case
frames created by the partial sentential structures as list
headings. In some cases the list item is a full syntactic
complement, in others the relation between item and heading is not
structural but only on semantic grounds.\\ Frequently, when
processing realistic documents deficiencies of the source material
have to be accounted for. An example: the analysis of the HTML
sources of foundry web pages revealed that HTML was sometimes
misused in the sense that the intended layout was not created by
the appropriate tagging (which would allow to easily recover the
intentions) but by misusing other tags to create a certain surface
appearance. Such deficient structures create a problem for
automatic analysis.\\
Examples of HTML misuse include:
\begin{itemize}
\item creation of a frame-like layout using tables, \item creation of
a list-like layout with $<$P$>$ and $<$BR$>$ tags (see example
\ref{ex_para}).
\end{itemize}

\begin{example} Misuse of Paragraphs. \label{ex_para}
\small
\begin{quote}
\flushleft{Folgende max. Abmessungen sind moeglich: $<$p$>$ - bis zu
14.000 mm Laenge$<$br$>$ - bis zu 6.000 mm Durchmesser $<$/p$>$}
\end{quote}
\end{example} \normalsize
\subsection{Tools and Resources}
In spite of the specific aspects discussed above analysis of textual
(parts of) web pages has a lot in common with document processing in
general. We therefore employ the XDOC\footnote{XDOC stands for
\emph{X}ML based \emph{doc}ument processing.} toolbox for this task
and do combine it with web page specific modules.
\subsection*{Methods}
For the analysis of information from web pages we need different
tools and resources. The tools can be divided into:
    \begin{itemize}
        \item Preprocessing tools like raw text extraction (`HTML cleaner')
        and collector of all web pages from a company resp. link tracing tools.
        \item Interpretation of HTML structures: what is the semantics behind HTML tags?
        \item Linguistic tools for the semantic interpretation of linguistic
        structures, like sentences or phrases.
    \end{itemize}
The \emph{WWW page preprocessing tools} are not directly relevant
for the analysis of implicit information, these tools only collect
and prepare the web pages. Relevant for the analysis of implicit
information is the interpretation of the \emph{internal structure
of web pages}: Which pieces of information are  embedded in which
HTML structures? Are they relevant for the semantic interpretation
of the contents inside the HTML structure?
The main focus is on the \emph{analysis of linguistic structures},
because inside HTML tags (e.g., tables or list structures etc.) we
can find continuous text as whole sentences, or on a lower level
phrases or simple lists of specific identifier, e.g., nouns or other
tokens. For this task we use our document suite XDOC -- a collection
of linguistic tools (see \cite{roesner.kunze:2002coling} for a
detailed description of the functions inside XDOC).
In all functions of XDOC the results of processing are encoded
with XML tags delimiting the respective piece of text. The
information conveyed by the tag name is enriched with XML
attributes
and their respective values.\\
In the following subsections we give a short description of
separate functions for the analysis of web pages. Examples of the
application of these functions are presented in section
\ref{sec_cast-nlp-examples}.
\begin{itemize}
%
  \item \emph{Part-of-Speech (POS) Tagger}: The assignment of part\--of\--speech information to a token -- POS
tagging for short -- is not only a preparatory step for parsing.
The information gained about a document by POS tagging and
evaluating its results is valuable in its own right.
We use a morphology based approach to POS tagging (cf.
\cite{roesner.kunze:2002coling} for details).
    \item \emph{Syntactic Parsing}: For syntactic parsing we apply a chart parser based on
    context-free grammar rules augmented with feature structures.
    \item \emph{Semantic Tagger}: For semantic tagging we apply a semantic lexicon. This lexicon
contains the semantic interpretation of a token and a case frame
combined with the syntactic valency requirements. Similar to POS
tagging, the tokens are annotated with their meaning and a
classification in semantic categories like, e.g., concepts and
relations.
    \item \emph{Case Frame Analysis}:
As a result of case frame analysis of a token we obtain details
about the type of recognized concepts (resolving multiple
interpretations) and possible relations to other concepts.
    \item \emph{Semantic Interpretation of Syntactic Structures (SISS)}:
Another step to analyze the relations between tokens can be the
interpretation of the specific syntactic structure of a phrase or
sentence. We exploit the syntactic structure of domain specific
sublanguages to uncover the semantic relations between related
tokens.
%
%
\end{itemize}
\subsection*{Resources}
The resources vary depending on the tools used (like grammars,
abbreviation lexicon etc.). For the analysis we need domain
resources, e.g., specific taxonomies of the domain, and document
specific resources. The document specific resources describe the
characteristics of the sublanguage inside the web pages (e.g., which
technical terms are used, what are the syntactic types of phrases
etc.).
\subsection{Web Page Classification} For an efficient processing of
web pages, which contain relevant information, a simple
pre-classification of web pages was developed:
\begin{itemize}
    \item {\it Information Pages:}\\
          Information pages contain continuous text.
          The information on these pages may be structured in tables or be given as mixed information
          in the form of continuous text and tables or
          numerations.
    \item {\it Lead Pages:}\\
          Lead pages contain links to other web pages about a single topic (or a single
          company). On these pages more links to internal web pages then to
          external web pages are found.
    \item {\it Overview Pages or Portals:}\\
          Overview Pages contain both text and links to external web pages.
          Here links to providers with a similar product range or to related topics are found.
\end{itemize}
This classification was chosen because of its applicability to other
domains. Web pages of other industrial sectors are organized in the
same manner, for example in producing industry and online dealers,
e.g., building industry (doors, windows, garages, real estate),
insurances, automotive industry. The classification of web pages into
information, lead and overview pages is sufficient for information
extraction with regard to the
creation of company profiles. \\
Parameters for an automatic classification of web pages are the
percentage of absolute text segments and hyperlinks of a web page,
for example, the number of internal links (e.g., one topic, company,
or domain), external links (e.g., other topics, companies or
domains), tokens, and pictures. In evaluating links local directory
structure is taken into account. Through specification of criteria
(e.g., number of internal links, the rate of internal and external
links, the rate of tokens and pictures etc.) the user may affect the
classification. Moreover web page classification is influenced by
layout based structures (frames, scripts, controlling elements, like
buttons).
\section{An Application Scenario}
This section is a case study about the creation of {\it casting
specific company profiles}.
There are about 300 German foundries and they are present in the
WWW with one or more web pages.
A possible scenario is the following: a product cannot completely
be produced in a company. In this case company profiles can be
used for choosing a supplier company. Companies use the WWW as a
kind of
`yellow pages' in order to get information about potential suppliers.\\
A company profile includes a variety of information about a specific
company. This information comprises product related data (e.g., size
and weight of the casting, material, moulding processes and quality
assurance resp. certificates) and company related data (e.g.,
single-piece work, small, middle, and mass production). In addition,
the location of the
potential supplier is important in order to reduce transportation costs.\\
Company profiles are applicable in two ways. First, existing
company web pages can be enhanced by making implicitly available
information explicit via semantic annotation. Second, the proposed
semantic annotation
of documents is usable in direct authoring of future company web pages.\\
\subsection{Creation of Company Profiles}
For the user it is important to know which products (e.g., boxes,
engine blocks, cylinder heads, or axles) for which industrial sectors
(e.g., motor industry, wind power industry, machine building industry
etc.) are produced by the company in question. This information
allows inferences with respect to quality requirements fulfilled by
the company during previous production processes. Company profiles
also contain data like
addresses and additional contact information.\\
About 60 foundry specific web pages have been analyzed with
respect to data required for company profiles. A first result is
the following XML-DTD: \small
\begin{example} DTD for a Foundry Company Profile. \label{ex_DTD}
\scriptsize
\begin{verbatim}
<!DOCTPYE profile [
 <!ELEMENT profile (foundry)+>
 <!ELEMENT foundry (name, specifics)>
 <!ELEMENT name (f_name, contact, address)>
   <!ELEMENT contact (tel, fax*, email*, http)>
   <!ELEMENT address (street, city, zip)>
   <!ELEMENT specifics (scope+, production, quality*)>
   <!ELEMENT scope (material, (weight*, dimension*))>
    <!ELEMENT dimension (#PCDATA)>
     <!ATTLIST dimension for_what (mould | dim) "mould">
   <!ELEMENT production (customer*, product*, i-sector*)>
   <!ELEMENT quality (#PCDATA)> ... ]>
\end{verbatim}
\end{example}
\label{ex_siss}
\normalsize The DTD is based on web page analyses and gives all
details of a company profile for foundries. Not all elements from
the DTD are used by each company profile. This kind of DTD is used
for presentation of contents. Parts of the DTD (e.g., address
information, measures) can be used for analysis of web pages with
our NLP tools.
\subsection{Instantiation of Elements}\label{sec_cast-nlp-examples}
In this section we present some examples for the recognition of
information which is needed for the creation of a  company
profile. As an illustrative case we again take profiles of casting
companies.\\
In casting a lot of measurements, like mould, weight etc. are
relevant.  For the detection of measurement information we use the
syntactic parser of XDOC. The grammar of the chart parser was
extended with rules, which describe structures like:
\texttt{\emph{2500 x 1400 x 600}} or \emph{1000 x 800 x 350 / 350
mm}. A general pattern for this can be described by the following
rules: \small
\begin{quote}
\textbf{MS-ENTRY}: number measure (optional) \\
\textbf{3D-MS-ENTRY}: MS-ENTRY x MS-ENTRY x MS-ENTRY
\end{quote}
\normalsize and for foundry specific dimensions: \small
\begin{quote}
\textbf{3D-MS-ENTRY-C}: MS-ENTRY x MS-ENTRY x
    MS-ENTRY / MS-ENTRY
\end{quote}
\normalsize Measuring units (like mm, kg etc.) are handled in a part
of the abbreviation lexicon.
Example \ref{ex_pos-tagger} shows the results of POS tagging of the
phrase: `\emph{Kastenformat 500 x 600 x 150 / 150 mm}'.
\begin{example} Results of POS Tagging.\label{ex_pos-tagger}
\scriptsize
\begin{verbatim}
<N>Kastenformat</N> <NR>500</NR> <ABBR>x</ABBR> <NR>600</NR>
<ABBR>x</ABBR> <NR>150</NR> <ABBR>/</ABBR> <NR>150</NR>
<ABBR>mm</ABBR>
\end{verbatim}
\normalsize
\end{example}
Numbers are tagged with \texttt{NR}, nouns with \texttt{N}. The
letter `x' -- used as a multiplication operator -- is in a first step
handled like an abbreviation (tag \texttt{ABBR}). In the future we
may work with a separate category for these specific symbols. With
the rules from above the syntactic parser interprets the results of
the POS tagger as the following structure:
\begin{example} Results of Syntactic Parsing.
\scriptsize
\begin{verbatim}
  <3D-MS-ENTRY-C RULE="MEAS4">
    <3D-MS-ENTRY RULE="MEAS3">
      <MS-ENTRY RULE="MEAS2"><NR>1000</NR></MS-ENTRY>
      <ABBR>x</ABBR>
      <MS-ENTRY RULE="MEAS2"><NR>800</NR></MS-ENTRY>
      <ABBR>x</ABBR>
      <MS-ENTRY RULE="MEAS2"><NR>350</NR></MS-ENTRY>
    </3D-MS-ENTRY>
    <ABBR>/</ABBR>
    <MS-ENTRY RULE="MEAS1"><NR>350</NR><ABBR>mm</ABBR>
    </MS-ENTRY></3D-MS-ENTRY-C>
\end{verbatim}
\normalsize
\end{example}
Now with the SISS approach we are able to assign the different
numbers in the phrase to height, length, width and the diameter
dimensions of a cylindric 3D-object. The SISS approach splits whole
syntactic structure into smaller structures and assigns a sense to
these structures. The possible assignments are coded in a lexicon.
For this example the assignments are shown in example \ref{ex_siss}:
\begin{example} Excerpt from the Lexicon for SISS. \label{ex_siss}
\scriptsize
\begin{verbatim}
<ASSIGNMENTS RULE="3D-MS-ENTRY-C">
    <COMPONENT NAME="3D-MS-ENTRY">
        <EXPAND>3-dimension</EXPAND></COMPONENT>
    <COMPONENT NAME="ABBR">NIL</COMPONENT>
    <COMPONENT NAME="MS-ENTRY">dimension-diameter</COMPONENT>
</ASSIGNMENTS>
<ASSIGNMENTS RULE="3D-MS-ENTRY">
    <COMPONENT NAME="MS-ENTRY">
        <EXPAND>dimension-height</EXPAND></COMPONENT>
    <COMPONENT NAME="ABBR">NIL</COMPONENT>
    <COMPONENT NAME="MS-ENTRY">
        <EXPAND>dimension-length</EXPAND></COMPONENT>
    <COMPONENT NAME="ABBR">NIL</COMPONENT>
    <COMPONENT NAME="MS-ENTRY">
        <EXPAND>dimension-width</EXPAND></COMPONENT>
</ASSIGNMENTS>
\end{verbatim}
\normalsize
\end{example}
For each child in a structure the assignments define an
interpretation; if the child structure is also a structure, which
is separately described in the lexicon, it will be annotated with
the tag \texttt{EXPAND}. The semantic sense is described through
the element of the tag \texttt{EXPAND}, e.g., dimension-length. A
\texttt{COMPONENT} with the element \texttt{NIL} means that this
child will not be interpreted.\\
Another method for the detection of semantic relations between
linguistic structures is XDOC's case frame analysis. A case frame
describes relations between various syntactic structures, like a
token (noun or verb) and its linguistic complements in the form of
noun phrases or prepositional phrases. An example: The german
phrase \hyphenation{Form-an-la-ge} `\emph{Formanlagen fuer
Grauguss}' can be semantically interpreted through the analysis of
the case frame of the token `\emph{Formanlage}' (see example
\ref{ex_case-frame}). The case frame contains a relation named
\texttt{TECHNIQUE}, the filler of this relation must be of the
semantic category \texttt{process} (described by the tag
\texttt{ASSIGN-TO}) and a syntactic structure of a prepositional
phrase with an accusative noun phrase and the preposition `fuer'
(tag \texttt{FORM)}. In our example the preposition phrase `fuer
Grauguss' is recognized as a filler for the relation
\texttt{TECHNIQUE}, because the token `Grauguss` is from the
semantic category \texttt{PROCESS}.
\begin{example} Results of Case Frame Analysis.
\label{ex_case-frame} \scriptsize
\begin{verbatim}
<CONCEPTS> <CONCEPT TYPE="tool">
    <WORD>Formanlagen</WORD>
    <DESC>Maschine</DESC>
    <SLOTS>
        <RELATION TYPE="TECHNIQUE">
            <ASSIGN-TO>PROCESS</ASSIGN_TO>
            <FORM>P(akk, fak, fuer)</FORM>
            <CONTENT>fuer Grauguss</CONTENT>
        </RELATION>
    </SLOTS>
</CONCEPT>
<CONCEPT TYPE="process">
    <WORD>Grauguss</WORD>
    <DESC>Fertigungsprozess</DESC>
</CONCEPT> </CONCEPTS>
\end{verbatim}
\normalsize
\end{example}
\section{Research Topics}
We report about work in progress. In the following we will sketch
some of the open questions to be investigated in the near future.
\paragraph{Treatment of
Coordination} To be applicable on a realistic scale, a toolbox for
document
    processing needs generic solutions on all levels of the linguistic
    system (lexical, syntactic, semantic, discourse).\\
    It is well known that there are always interactions and
    inter\-dependencies between different levels. An example: In our
    current work we investigate possible solutions for the issues of
    coordinated structures.  These structures need proper treatment on
    the lexical level (e.g.,  treatment of prefix and suffix truncation
    in POS tagging), syntactic level (e.g., grammar rules for adjective
    groups, noun groups and mixtures of both) and on the semantic
    level (e.g., rules to decide about a disjunctive vs. a conjunctive
    reading).\\
    Complex coordinated structures are relevant in virtually all our
    technical and medical applications. Thus our solutions should
    clearly be generic and independent of the domain, but domain
    knowledge as a resource may be needed for semantic
    interpretation.\\
    Coordinated structures are relevant for querying documents as
    well. As a point in case see phrases like {\it Klein- und
    Mittelserien} (in english: small batch and middle production).  If
    you search for {\it Kleinserien} with the help of a conventional
    search engine, you probably will not find web pages with the
    phrase {\it Klein- und Mittelserien}.
\paragraph{Towards Generic Solutions} NLP has both an analytic and a
generative `procedural' perspective in the sense that grammars are
not only used to merely `describe' linguistic structures but to
actually analyze or generate them. The analogy for documents would be
to not only use descriptions (DTDs, XML Schemata) to validate already
marked up instances but to employ descriptions both for analysis of
yet unmarked documents as well as for the generation of documents
that obey the rules of a
schema.\\
The move from DTDs to XML Schemata is a big step forward and allows
to better model document contents and structure. From the perspective
of not only describing but automatically analyzing documents further
improvements should be envisaged. Will it be possible to integrate
information about the automatic detection of document elements into
an extended document description framework? Such a possibility would
allow to declaratively configure schemata for documents that could
then be exploited for the processing of raw documents that implicitly
follow the schema but do not have explicit markup yet.
\paragraph{Limits of Markup}
Where are the limits for enriching documents with XML markup? The
basic idea of delimiting spans of text in a document within an
opening and a closing tag is simple and convincing. But: will simple
well formed markup really suffice? When will we unavoidably have to
work with concurrent markup \cite{concurrent}? When will we have to
give up the adjacency requirement for text spans and will have to
deal with discontinuous structures as well?

\section{Conclusion}
The vision of the Semantic Web is a stimulating driving force for
research in text technology, Computational Linguistics (CL) and
knowledge representation. The basic issues are actually not new, but
they now receive a much broader attention than before.
These different approaches are complementary, combining them will
result in synergetic effects.\\
The focus of the SGML/XML approaches to documents has been on
providing means to describe document structures (i.e., DTDs,
schemata) and on tools to validate already marked up document
instances with respect to an abstract description (i.e., parsers,
validators).  From an NLP perspective the weakest point in
SGML/XML is that the contents of terminal elements (i.e., elements
with text only) is simply uninterpreted PCDATA, i.e., strings
without their internal structure made explicit. Although in
principle markup can go down to the granularity of single
characters there is a `traditional' bias in the markup community
towards macro
level structuring with paragraphs as the typical grain size.\\
On the other hand, many -- but definitely not all -- techniques
and tools from CL and NLP have their focus on the word, phrase or
sentence level. Why not combine macro level structuring with
lexical, syntactic and semantic analysis of otherwise
uninterpreted PCDATA?\\
The third field -- knowledge representation -- is indispensable
because semantics needs grounding in formal KR structures.
Ontologies (i.e., basic vocabularies for expressing meaningful
relations) and worked out a taxonomy of interrelated domain
concepts are the backbone of any semantic account of document
contents.\\
\section*{Acknowledgement} We have to thank the anonymous reviewers
for their comments and our colleagues, J. Kapfer and M. Piotrowski,
for their efforts in thorough proofreading and their constructive
criticism.
\bibliographystyle{plain}
\bibliography{final}
\end{document}